\documentclass[12pt,letterpaper]{article}
\usepackage[top=0.85in,left=0.75in,footskip=0.75in,marginparwidth=2in]{geometry}

\usepackage[utf8]{inputenc}

\usepackage[backend=bibtex,style=numeric-comp,natbib=true,url=false]{biblatex}
\AtEveryBibitem{%
  \clearfield{issn} 
  \clearfield{doi} 
  \clearfield{note} 

  \ifentrytype{online}{}{
    \clearfield{url}
  }
}
\usepackage{nameref,hyperref}

\usepackage[right]{lineno}

\usepackage{microtype}
\DisableLigatures[f]{encoding = *, family = * }

\raggedright
\usepackage{changepage}

\usepackage[aboveskip=1pt,labelfont=bf,labelsep=period,singlelinecheck=off]{caption}

\makeatletter
\renewcommand{\@biblabel}[1]{\quad#1.}
\makeatother

\usepackage{lastpage,fancyhdr,graphicx}
\usepackage{epstopdf}
\fancyheadoffset{0.4in}
\fancyfootoffset{0.4in}

\usepackage{color}

\definecolor{Gray}{gray}{.25}

\usepackage{graphicx}

\usepackage{sidecap}

\usepackage{wrapfig}
\usepackage[pscoord]{eso-pic}
\usepackage[fulladjust]{marginnote}
\reversemarginpar

\usepackage{algorithm}
\usepackage[noend]{algpseudocode}

\usepackage[space]{grffile}

\def\bw{\mathbf{w}}
\def\bx{\mathbf{x}}

\usepackage{amsmath,amssymb}

\newcommand{\Exp}[1]{\left\langle\,{#1}\,\right\rangle}

\newcommand{\appropto}{\mathrel{\vcenter{
  \offinterlineskip\halign{\hfil$##$\cr
    \propto\cr\noalign{\kern2pt}\sim\cr\noalign{\kern-2pt}}}}}

\begin{document}
\vspace*{0.05in}

\begin{flushleft}
{\Large
\textbf\newline{Scaling of learning time for high dimensional inputs}
}
\newline
\\
Carlos Stein Brito
 \\
 \bigskip
 NightCity Labs, Lisbon, Portugal
\end{flushleft}

\justify

\section*{Abstract}
Representation learning from complex data typically involves models with a large number of parameters, which in turn require large amounts of data samples. In neural network models, model complexity grows with the number of inputs to each neuron, with a trade-off between model expressivity and learning time. A precise characterization of this trade-off would help explain the connectivity and learning times observed in artificial and biological networks. We present a theoretical analysis of how learning time depends on input dimensionality for a Hebbian learning model performing independent component analysis.
Based on the geometry of high-dimensional spaces, we show that the learning dynamics reduce to a unidimensional problem, with learning times dependent only on initial conditions. 
For higher input dimensions, initial parameters have smaller learning gradients and larger learning times. We find that learning times have supralinear scaling, becoming quickly prohibitive for high input dimensions. These results reveal a fundamental limitation for learning in high dimensions and help elucidate how the optimal design of neural networks depends on data complexity. Our approach outlines a new framework for analyzing learning dynamics and model complexity in neural network models.



\section{Introduction}

Neural networks demand large amounts of data and learning time to be trained \citep{krizhevsky_imagenet_2012}. Recent breakthroughs in machine learning owe much of their success to the increasing availability of computing power and large-scale datasets \citep{lecun_deep_2015}.
Even with such large resources, learning time remains one of the main obstacles in scaling up the complexity of neural networks and their applications \citep{le_building_2013}.

Both the number of neurons and the number of inputs per neuron (fan-in) influence learning time. Also biological neural networks, with billions of neurons and thousands of synapses contacting each of them \citep{kandel_principles_2000}, are likely to be constrained by similar trade-offs. Here we analyze the learning dynamics in neural networks and develop insights into how network complexity affects learning time.

We consider the unsupervised learning task of finding sparse hidden features in $N$ dimensional data. This task can be solved by a single neuron with a nonlinear Hebbian learning rule \citep{oja_learning_1991}. We show that the optimization function of the problem has a number of saddle points and maxima that increase exponentially with the dimensionality. We show that for a large number $N$ of input connections, it becomes highly probable that random initial synaptic weights will be almost orthogonal to the hidden features, in the regions of saddle points, which leads to large learning times. 

 Based on the geometry and statistical properties of high-dimensional spaces, we show that the learning dynamics are well described by a single dynamical system, in which the only relevant variable is the initial condition, determined by the distance to the closest hidden feature. We show that this initial overlap decreases with the number of inputs, which leads to a rapid increase in learning time for larger dimensions. 

 Our results lead to a surprising new finding: the optimal learning time has a supralinear dependency on the number of synaptic weights onto a single neuron. This dependency induces a fundamental limitation on synaptic connectivity, as learning becomes obstructively slow for a large number of inputs. We expect this approach to be useful for understanding other networks and tasks, for instance, convolutional network models in computer vision models. More generally, our approach outlines a new framework for analyzing learning dynamics and model complexity in neural network models.

\section{Results}

\subsection{Nonlinear Hebbian learning of sparse features}

We consider the projection pursuit problem of finding hidden sparse features in the neuron's $N$ inputs, as in independent component analysis settings. We search for the hidden patterns by optimizing the synaptic weights $\bw$ according to an objective function, 
\begin{equation}
\text{max}_{\bw, ||\bw|| = 1} \Exp{F(\bw^T \bx)}
\label{eq:ppdefbound}
\end{equation}
We model the $N$-dimensional input as a linear combination of $K$ sparse hidden variables. We constrain the norm of the weights to $|\bw|_2 = 1$, optimizing only the direction of $\bw$, and we use whitened inputs, $\langle \bx \bx^T \rangle = I$. In general, for this task, the specific shape of the sparse distributions is not important \citep{oja_learning_1991, brito_nonlinear_2016}, and we consider two cases: symmetric (Laplacian) or asymmetric distribution ($\chi^2$). We develop here our analysis for the symmetric case.

Many possible optimization functions $F(.)$ are appropriate for learning sparse components \citep{oja_learning_1991, brito_nonlinear_2016}. Learning this objective by stochastic gradient descent leads to the nonlinear Hebbian learning rule, $\Delta \bw_t \propto \bx_t \ f(\bw^T \bx_t)$, where $f(u) = \frac{\partial F}{\partial u}$. We use a linear rectifier in our simulations, $f(u) = (u-2.)_+$.



\subsection{The geometry of the optimization surface for input weights}

\begin{figure}[!htb]
\begin{center}
 \includegraphics[width=\textwidth]{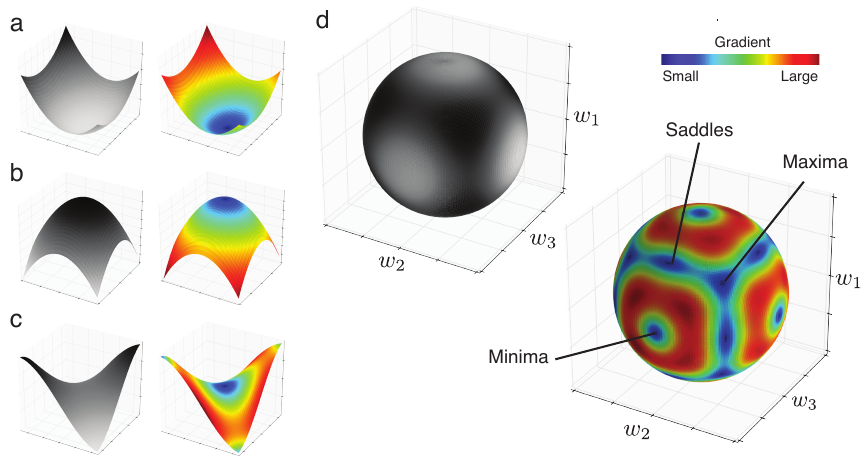}
\end{center}
\caption[Geometry of the optimization surface for synaptic weights.]{\textbf{Geometry of the optimization surface for synaptic weights.} (\textbf{a}) Prototypical minimum in a convex surface with a basin of attraction. (left) Gray scale indicates optimization function value, with minimum in white. (right) Color heat map indicates the magnitude of the gradient, with zero gradients in dark blue. (\textbf{b}) Prototypical maximum in a concave surface, with an unstable equilibrium point. (\textbf{c}) Saddle points are concave in some parameter directions, while convex in others. (\textbf{d}) Surface for the gradient magnitude for three weights. Each dimension represents the value of a synaptic weight. As we enforce $|\bw|_2 = 1$, the parameter space is constrained to the unit sphere. The cardinal directions are the minima, the directions of the hidden patterns. The symmetric directions, where all weights have the same magnitude, are maxima. Partially symmetric directions, where two weights have the same magnitude, and the third is zero, are saddle points. Areas in blue indicate low gradient magnitude, and thus slow learning dynamics.}
\label{fig:bounds:3d}
\end{figure}

To understand the learning dynamics of our model, we start by developing an analytical description of the optimization surface. We determine the position and quantity of minima, maxima, and saddle points, which are the critical points for gradient descent methods. Saddle-points have zero derivatives, but present positive curvature in some parameter directions, while negative in others. Figure \ref{fig:bounds:3d}a-c illustrates the typical geometry of these special points on an optimization surface in two dimensions.


Given the input distribution with $K = N$, with one hidden feature at each cardinal direction, the optimization function has a minimum $\bw^*$ at each of the $N$ hidden features, $\bw^* = (0,\dots,0,\pm1,0,\dots,0)$, yielding a total of $2 N$ minima. At the same time, there are exponentially many maxima, at each of the $2^N$ symmetric directions, $\bw^{max} =  (\pm1,\pm1,\dots,\pm1,\pm1)/\sqrt{N}$.


In addition, there exists an even larger number of saddle points. Although hard to visualize, any direction with $k$ symmetric components ($1 < k < N$), $|w_{i_1}| = ... = |w_{i_k}| \neq 0 $, while other components are zero, is a saddle point. It is a maximum in respect to the $k$ non-zero components, and a minimum in respect to the other $N-k$ null components. We show in Methods that it implies a total number on the order of $3^N$ saddle points.

Figure \ref{fig:bounds:3d}d illustrates these properties for the three-dimensional case, $N=3$. The 6 cardinal directions represent the minima, where the hidden features lie. Each one of these is surrounded by a basin of attraction. The 8 symmetric directions are maxima, while the 12 partially symmetric points are the saddle points. 
 
The parameter regions that connect maxima and saddle points have small gradients and are far from the minima. Starting at these regions leads to large learning times. Note that the maxima and saddles exist due to the symmetries in the problem. Saddle points due to symmetries have previously been observed in other network models \citep{saxe_exact_2013,choromanska_loss_2014,dauphin_identifying_2014}. The numerical predominance of saddle points is also in alignment with these other studies \citep{saxe_exact_2013,choromanska_loss_2014,dauphin_identifying_2014}. In contrast to these studies, where the symmetries are due to the invariance to permutations of neurons inside a network layer, out symmetries are for weights of a single neuron.


\subsection{Quasi-orthogonal random directions in high-dimensional spaces}

The three-dimensional example illustrated above gives a good intuition for some general properties of the optimization surface of feature learning by projection pursuit. However, high-dimensional spaces have anti-intuitive geometrical properties that are not evident in low-dimensional cases. The essential property that we will exploit for our analysis is the angular distance between random vectors, as the weight initial values are usually randomly distributed.

 Before we turn to random vectors, let us consider a "worst-case" scenario.
In the three dimensional case, any vector will be at most at $55^{\circ}$ angular distance to one of the cardinal directions, with the maximal distance at the maxima (worst case). However, in higher dimensions this distance increases asymptotically to $90^{\circ}$, with the overlap between minima and maxima in $N$ dimensions being $d_{\text{max}} = \bw_{\text{max}}^T \bw^* = 1/\sqrt{N}$ (see Methods).

\begin{figure}[!htb]
\begin{center}
 \includegraphics[width=0.9\textwidth]{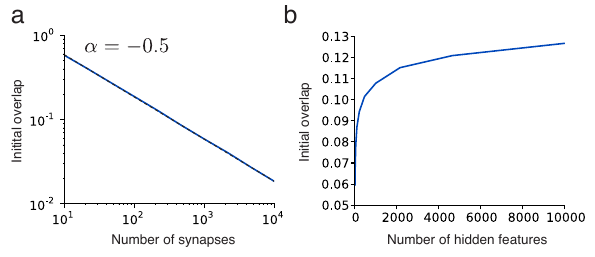}
\end{center}
\caption[Initial parameters have small overlap with hidden features in high dimensions.]{\textbf{Initial parameters have small overlap with hidden features in high dimensions.} (\textbf{a}) The expected overlap of initial parameters with a hidden feature decays with $\sqrt{N}$ (fixed $K = 10$). Dashed line is power-law with exponent $\alpha = -0.5$. (\textbf{b}) Increasing the number of hidden features has only a logarithmic effect on the expected initial overlap (fixed $N = 1000$).}
\label{fig:dim:initial}
\end{figure}

Random directions $\bw^R$ follow a similar decay in their average distance $d$ to the closest minimum, $d \approx \frac{\sqrt{2 \ \text{log}(N)}}{\sqrt{N}}$ (see Methods). This result implies that initial parameters will have only a small overlap with the hidden features, starting from almost orthogonal angles. Although we have so far assumed the same number of hidden features $K$ and input dimensions, $K = N$, our results can also be adapted to $K < N$ and $K > N$ (see Methods), leading to an overlap given by
\begin{equation}
d^K \approx \frac{\sqrt{2 \ \text{log}(K)}}{\sqrt{N}}
\label{eq:1n}
\end{equation}
Thus, for large $N$, the expected overlap will still be small even if when the number of hidden features is large in relation to the number of dimensions ($K > N$).
It is a by-product of the fact that in high dimensions one may generate an exponential number of random vectors, and they will be quasi-orthogonal to each other with high probability \citep{cai_distributions_2013}. In Figure \ref{fig:dim:initial} we illustrate these dependencies in a simulation with randomly generated directions. Figure \ref{fig:dim:initial}b shows how, given an input dimensionality $N$, the initial overlap $d^K$ increases only logarithmically with the number of hidden features.

We can relate these facts to the intuition gained in the three-dimensional case.  When the number of dimensions increases, the area of the blue region filled with saddles and maxima (see Figure \ref{fig:bounds:3d}d) becomes exponentially larger than the area composed by the basins of attraction around the minima. Formally, with higher dimensions, the area of the polar cap around a given direction decreases exponentially in comparison to the total area of the sphere \citep{li_concise_2011}. 

These results imply that when there are many input weights, the initial random weights will be quasi-orthogonal to hidden features, and lie in the parameter region of small gradients that is filled with saddle points and maxima.

\subsection{Reduction to unidimensional dynamics}

With the general landscape of the optimization problem in hand, we now analyze the gradient near the critical points and initial conditions. We assume here large dimensionality $N \gg 1$ and $K = N$, with initial parameters having a small overlap with the hidden features, $d \to 0$. 

In the projection pursuit problem, the total input to the neuron is a mixture of the $N$ hidden patterns, weighted by the synaptic strengths, $u = \bw^T \bx = w_1 x_1 + \dots + w_N x_N$. Since for large $N$ the initial individual weights $w^0_i$ are small, we may invoke the central limit theorem, and conclude that, as $N$ increases, the total input $u(\bx)$ converges to a normal distribution.

We make the assumption that during learning the weights will converge to the closest hidden pattern $\bw^* = \mathbf{e}_j = (0, \dots , 0, 1, 0, \dots , 0)$, following the shortest path from the initial weights $\bw^0 = (w_1^0, \dots , w_{j-1}^0, w_{j}^0 = d_0, w_{j+1}^0, \dots , w_N^0)$, where $j$ is the feature with maximal overlap and $d_0$ is the initial overlap. 

We rewrite the input as $u = w_j x_j + \sum_{j \neq k} w_k \ x_k$, and invoke again the central limit theorem, arriving at the approximation $u = w_j x_j + \sum_{j \neq k} w_k \ g = w_j \ l + \sqrt{1 - w_j^2} \ g$, where $l$ is a Laplacian variable and $g$ is a normally distributed variable, and we used the fact that the weights are normalized, $\sum w_j^2 = 1$.

Thus along the path from $\bw^0$ to $\bw^*$ we can reduce the change in $u$ to a linear transition between the initial (Gaussian) distribution and the final (sparse) distribution, 
\begin{equation}
u_t = d_t \ l + \sqrt{1 - d_t{^2}} \ g
\label{eq:utd}
\end{equation}
parameterized by the overlap $d$.

The formulation in Eq. \ref{eq:utd} reveals that the distribution of inputs $u_t$ depends only on the overlap $d_t$. It implies that the gradient of $\Exp{F(u)}$ depends only on $d_t$, allowing us to study the learning dynamics through a single onedimensional system. With $\hat{F}(d_t) = \Exp{F(u_t)}$, where $u_t$ is given by Eq. \ref{eq:utd}, gradient descent gives
\begin{equation}
\Delta d_t \propto \frac{\partial \hat{F}(d_t)}{\partial d}
\label{eq:unid}
\end{equation}

\begin{figure}[!htb]
\begin{center}
 \includegraphics[width=0.9\textwidth]{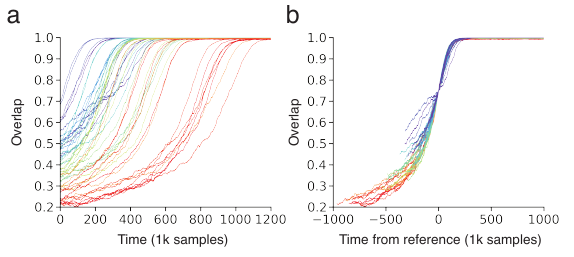}
\end{center}
\caption[Stereotypical learning dynamics in high-dimensional optimization.]{\textbf{Stereotypical learning dynamics in high-dimensional optimization.} (\textbf{a}) Evolution of the overlap of the weights with a hidden feature, for different dimensionality $N$. The color heat indicates $N$, varying from $N = 10$ (purple) to $N = 160$ (red). One run for each $N$, with random initial weights. (\textbf{b}) Same trajectories shifted to a referent time where an overlap of $d = 0.75$ was reached, highlighting the similarity in the learning dynamics.}
\label{fig:dim:traj}
\end{figure}

We illustrate in Fig. \ref{fig:dim:traj} how the full $N$-dimensional learning dynamics can be represented by the effective one-dimensional dynamics (Eq. \ref{eq:unid}) for different dimensionality $N$. Figure \ref{fig:dim:traj}a shows the time evolution of the largest overlap $d_t$ between $\bw_t$ and a hidden feature. It shows that initial overlaps depend on dimensionality, with higher dimensions with lower initial overlap on average, and thus larger learning time. 

It also illustrates how all runs have similar trajectory profiles, as expected from a reduction to a one-dimensional system. In Figure \ref{fig:dim:traj}b we highlight the stereotypical learning dynamics by showing the trajectories with a shifted time reference, to the point when the overlap crossed $d_t = 0.75$.

\begin{figure}[!htb]
\begin{center}
 \includegraphics[width=0.7\textwidth]{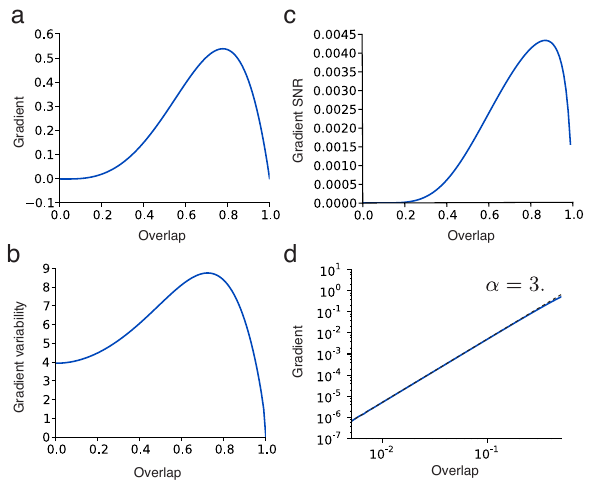}
\end{center}
\caption[Gradient dependency on the overlap.]{\textbf{Gradient dependency on the overlap $d$ for symmetric sparse distributions.} (\textbf{a}) The gradient magnitude $\mu(d)$ vanishes when the overlap goes to zero. (\textbf{b}) The gradient variability $\sigma(d)$ does not change significantly for small overlaps. (\textbf{c}) The gradient signal-to-noise ratio $\mu(d)^2/\sigma(d)^2$ follows the profile of the gradient magnitude. (\textbf{d}) For small overlaps, $d \to 0$, the gradient has a power-law dependency on the overlap (dashed line is power-law with exponent $\alpha = 3.$).} 
\label{fig:dim:grad}
\end{figure}

We can analyze the properties of the gradient $\mu(d) = \frac{\partial \hat{F}(d_t)}{\partial d}$ for small overlaps. 
We Taylor expand the objective $\hat{F}(\textbf{d})$ around $d=0$ and, for symmetric feature distributions, the first three moments are zero, leading to 
\begin{equation}
 \hat{F}(d) \propto a + d^4 \implies  \mu(d) \propto d^3
\label{eq:mud}
\end{equation}
In Figure \ref{fig:dim:grad}d, we confirm this scaling law by calculating the gradient for small overlaps in simulations. Asymmetric sparse distributions have a non-zero third moment, leading to a $\mu(d) \propto d^2$ scaling. 

\subsection{Learning time dependence on input dimensionality}

To estimate the learning time, we consider the optimal learning rate $\eta$ at a given overlap $d$. After a learning and normalization step, we have (see Methods):
\begin{align}
\Delta d  & \approx \eta \mu - \frac{1}{2} \eta^2 N \sigma^2 d
 \label{eq:sgdstept} 
\end{align}
with a positive term given by the gradient $\mu$, and a negative term that scales with $N$ due to the noise of the gradient $\sigma$ in all weight dimensions. We derive the optimal learning rate, and optimal change in $d$,
\begin{align}
\eta^* = \frac{2 \mu}{N \sigma^2 d}  \implies \Delta d  & \approx \frac{2}{N d} \frac{\mu^2}{\sigma^2} 
\label{eq:etaopttext} 
\end{align}
where we see a dependence on the signal-to-noise ratio of the gradient. In Figure \ref{fig:dim:grad} we show the gradient statistics dependent on the overlap $d$, by simulating our reduced unidimensional model of Equation \ref{eq:unid}. 
We see that the gradient SNR goes quickly to zero for small overlaps (Figure \ref{fig:dim:grad}c). 
Also, the SNR profile follows the profile of the gradient magnitude $\mu$ (Figure \ref{fig:dim:grad}a), as the gradient variability $\sigma$ is approximately constant for small overlaps (Figure \ref{fig:dim:grad}b).

We can now estimate the total learning time based on the change $\Delta d$ for optimal learning rates, from a starting point $d_0$ to a target $d^*$,
\begin{equation}
T(d_0) \approx \int_{d_0}^{d^*} \frac{N d}{2} \frac{\sigma^2}{\mu^2} \partial \tilde{d}
\label{eq:tmu}
\end{equation}





We have succeeded to derive a chain of dependencies, in Eqs. (\ref{eq:1n}), (\ref{eq:tmu}) and (\ref{eq:mud}), that together reveal that the learning time has a supralinear dependency on the input dimensionality. For the symmetric case, we have
\begin{equation}
\left.
\begin{aligned}
d_0 & \propto \sqrt{log(K)/N} \\
\mu & \propto d^3 \\
T & \propto \int_{d_0}  \frac{N d}{2} \frac{\sigma^2}{\mu^2} \ \partial d
\end{aligned}
\ \
\right\}
\implies T \propto \frac{N^{3}}{log(K)^2}
\end{equation}
while for the asymmetric case, with $\mu \propto d^2$, we have
\begin{align}
T \propto \frac{N^{2}}{log(K)}
\end{align}

\begin{figure}[!htb]
\begin{center}
 \includegraphics[width=0.5\textwidth]{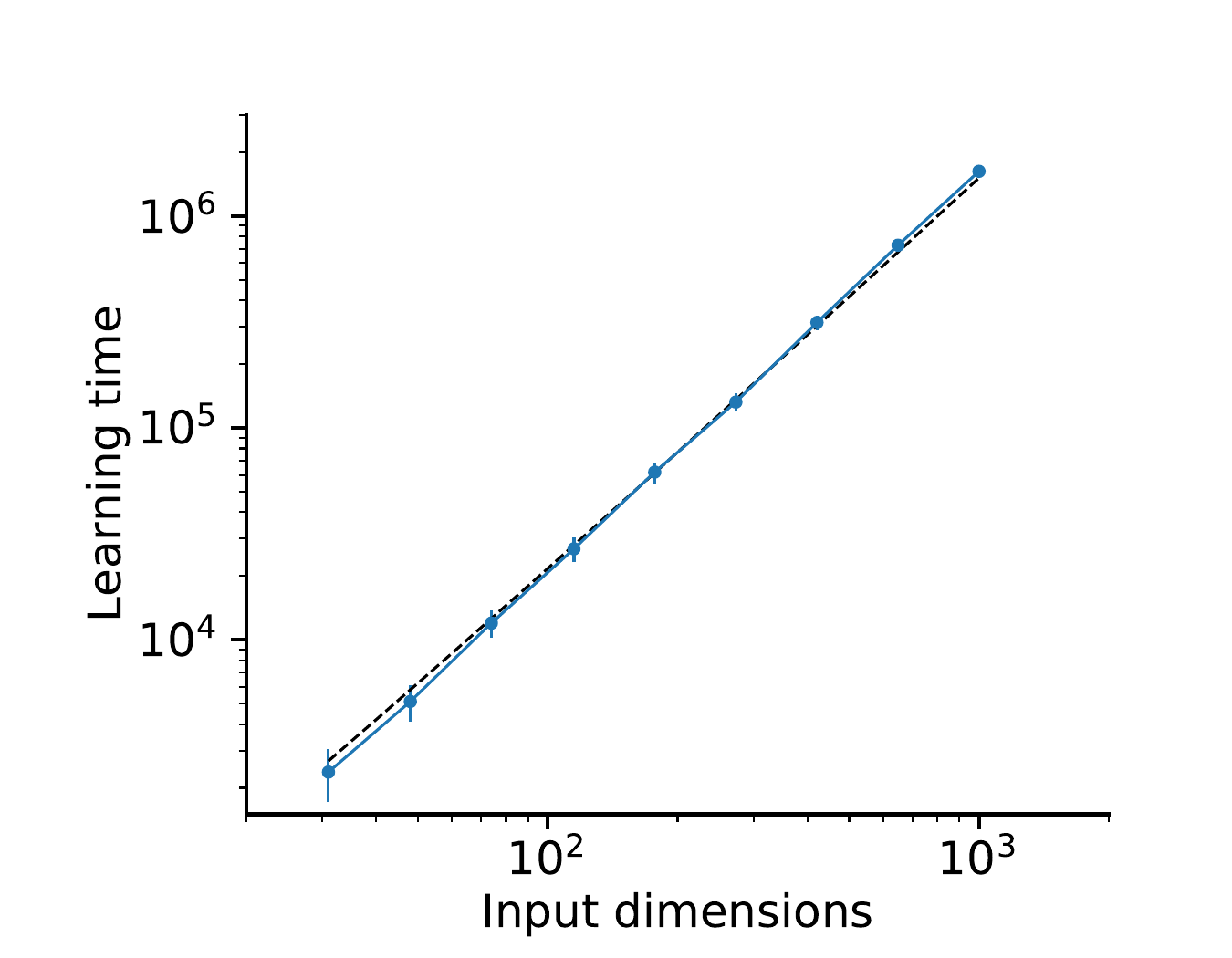}
\end{center}
\caption[Learning time dependence on the number of inputs.]{\textbf{Learning time dependence on the number of inputs.} Learning time for input dimensions $N$, and $K=N$ latent features (in blue) match the predicted scaling  by the theory, $T \propto \frac{N^{2}}{log(N)}$. Learning time was defined as the passing time of the overlap at $d = 0.7$, averaged over 100 simulations.}
\label{fig:dim:tvsn}
\end{figure}

In Figure \ref{fig:dim:tvsn}, we show for the asymmetric case that our gradient descent simulations have a learning time that scales with $\frac{N^{2}}{log(K)}$.



\section{Discussion}



\subsection{Extensions to alternative neural network models}

We have studied a particular model of unsupervised learning. However, it is possible that our results can be extended to other learning models. 
For any network model in which the total input to the neuron is a linear projection, $u = \bw^T \bx$, the gradient for the synaptic strengths can be formalized as $g = \partial_w \Exp{F(\bw^T \bx)}_x$, as in our study. Reinforcement learning or supervised learning networks are important examples.

The crucial assumption made for our unsupervised learning scenario is that the optimization function only depends on higher-order statistics of the input. Future investigations shall determine if we can extend the approach to alternative learning paradigms, where target values or rewards also play a role.

\subsection{Theory for convolutional neural networks}

Large hierarchical neural network models have recently obtained impressive results in a variety of artificial intelligence tasks \citep{lecun_deep_2015,bengio_representation_2013}. A large part of these applications relies on the implementation of convolutional neural networks, in which each neuron has a limited receptive field size \citep{krizhevsky_imagenet_2012,mnih_playing_2013}. This architectural constraint is essential when the input has high dimensionality, as in the case of images. 

Despite its popularity and proven efficiency, a definite theoretical explanation for the functional gains of limited input dimensions has been elusive. Our results suggest a potential explanation for the performance gains due to localized receptive fields and connectivity constraints, showing analytically that larger receptive field sizes can make learning impractical. In follow-up work, we are investigating if the proposed scaling laws apply to large-scale convolutional networks.

\subsection{Implications for neural network learning dynamics}

The unsupervised learning paradigm we have used has proven to be a useful prototype in understanding how learning unfolds in neural networks. It allows us to calculate precisely the number of extreme points, such as saddle points, how they are distributed and how they determine the learning dynamics.

Our curvature analysis is aligned with previous studies that have collected evidence that saddle points are abundant and may be an important obstacle for neural network learning \citep{saxe_exact_2013,dauphin_identifying_2014,choromanska_loss_2014}. In \citet{dauphin_identifying_2014}, the authors use numerical simulations to probe the existence of saddle points in large networks. Our results may help to provide an analytical explanation for their findings. 

In \citet{saxe_exact_2013}, the authors study a multi-layer linear network model, where they use the optimization gradient to qualitatively characterize the learning dynamics. However, the choice of a linear model led to different conclusions, and their results depend on properties of multi-layer architectures, but not on the learning properties of each neuron or the number of synapses. In their model, a single-layer network does not have saddle points, and the symmetries in the optimization were due to the multiple layers, providing a complementary source of symmetries in large neural networks. Our analysis allows the study of nonlinear multi-layer networks, and in ongoing work, we investigate how our results generalize to multiple layers.



\subsection{Theory for localized receptive fields and number of synapses in the cortex}

Cortical neurons have in the order of $1000$ to $10000$ input synapses and as space is a strong constraint in the brain, it is believed connectivity properties may be determined by the trade-off between representational power and volume \citep{rivera-alba_wiring_2011}. 

Our results provide an alternative trade-off for synaptic densities. Synaptic plasticity of pyramidal neurons has Hebbian mechanisms that can be mapped into the nonlinear Hebbian learning rule studied here \citep{pfister_triplets_2006,brito_nonlinear_2016}. For such models, we have shown that learning is effectively limited by the number of input synapses per neuron, and even in the absence of physical constraints, there would still be a limit on the order of thousands of synapses imposed by the learning time trade-off. 



Our theory also makes predictions about the time scales and learning dynamics in sensory development. 
Given the statistics of the neural activities, and models of synaptic plasticity, we may use our theory to estimate how much sensory data is needed for learning. It would provide a theoretically grounded explanation for the time scales of critical periods in sensory development \citep{berardi_critical_2000}.



\section{Methods}


\subsection{Number of maxima and saddle points}

We provide here a short explanation for how to count the maxima and minima in the synaptic optimization surface for symmetric latent features. Maxima lie at the edges of a hypercube, where all weights have the same magnitude, $w_i = \pm1 / \sqrt{N}$. For each weight there are two possibilities, negative or positive, amounting to $2^N$ combinations.

Saddles have some zero weights, while the others have the same magnitude, $|w_{i_1}| = \dots = |w_{i_k}| \neq 0$. Thus each weight has three possibilities, positive, negative or zero, amounting to $3^N$ combinations. The exceptions are the points that are minima, maxima, and the zero vector $\bw = 0$, totaling $3^N - 2^N - 2 N - 1$ saddle points.

\subsection{Data generation}

In general, the $N$-dimensional input $\bx$ to our unsupervised learning problem is generated by a whitened linear mixture of $K$ non-Gaussian variables, $\tilde{\bx} = \sum_{i=1}^K \bw_i l_i$, where is a normalized Laplacian variable in the symmetric case ($l_i \sim \text{Laplace}(0,1/\sqrt{2})$) or a normalized $\chi^2$ variable for the asymmetric case ($l_i \sim \chi^2_q(-q,1/\sqrt{2q})$, with $q=10$), and the mixing vectors $\bw_i$ are $N$-dimensional. 

For $K = N$, we used orthogonal mixing vectors, $\bw_i = (0, \dots, 0, w_i = 1, 0, \dots, 0)$. For $K < N$, we add Gaussian variables to complete $N$ dimensions, generating the input by $\bx = \sum_{i=1}^K \bw_i l_i + \sum_{i=K+1}^N \bw_i g_i$, where $g_i \sim \text{Normal}(0,1)$, and same mixing vectors as in the previous case. For $K > N$, we generate the input by $\tilde{\bx} = \sum_{i=1}^K \bw_i l_i$, for random vectors $\bw_i$ with $|\bw_i| = 1$. The resulting data is then whitened, $\bx = \mathbf{M} \ \tilde{\bx}$.

\subsection{Expected overlap between random vectors in high dimensional spaces}

The average overlap between two random directions can be derived from how random directions $\bw^R$ are generated in an $N$-dimensional sphere: each component is drawn from a normally distributed variable, and the resulting vector is normalized to unit norm,
\begin{align}
\tilde{\bw} & = (\tilde{w}_1, \dots, \tilde{w}_N), \ w_i \sim N(0,1) \\
\bw^R & = \frac{\tilde{\bw}}{|\tilde{\bw}|}
\end{align}
Since for large $N$ the norm $|\tilde{\bw}|$ is well approximated by $\sqrt{N}$, we conclude that each component of the random vector follows $w_i^R \sim N(0,\frac{1}{\sqrt{N}})$. 
If we consider without loss of generality that the reference direction is $\bw^* = (1,0,\dots,0)$, the overlap will follow a distribution $d \sim N(0,\frac{1}{\sqrt{N}})$.

Considering $K$ random directions $\bw^k$ instead, the expected largest overlap $d^K$ to a reference direction, $\bw^* = (1,0,\dots,0)$, is given by the maximal value amongst the $K$ first components, $d^K = max_k \ w^k_1 \approx \frac{\sqrt{2 \text{log}(K)}}{\sqrt{N}}$, where we used the expected extreme value amongst $K$ normally distributed variables for large $K$, $max_k \ g_k \approx \sqrt{2 log K}$ \citep{david_order_1970}. 

It follows that if there are $K$ random hidden features in an $N$-dimensional space, the initial random parameters will have an expected largest overlap of approximately $\frac{\sqrt{2 \text{log}(K)}}{\sqrt{N}}$ to one of the hidden features.


\subsection{Gradient scaling law for small overlaps}

We analyze the gradient magnitude for a small overlap between the weights and the hidden feature, $d \to 0$. 
We consider the statistical cumulants of the projected input $u$, where $u = d \ l + \sqrt{1 - d^2} \ g$. As the variance is constant, the second cumulant is constant, $\kappa^u_2 = 1$. Higher cumulants depend only on $l$, $\kappa^u_{m} = \kappa^{l}_{m} d^{m}$, for $m > 2$, since the normal variable $g$ has null higher order cumulants, $\kappa^{g}_{m} = 0, m > 2$. 

We divide the analysis into two cases, for symmetric and asymmetric distributions for the hidden feature $l$. For symmetric distributions, the odd cumulants $\kappa_{2m+1}^u$ are zero. We calculate the Taylor expansion of the optimization function at $u = 0$,
\begin{align}
\langle F(u) \rangle = & \ \langle F(0) + u F^{(1)}(0) + u^2 F^{(2)}(0)/2 + u^3 F^{(3)}(0)/3! + u^4 F^{(4)}(0)/4! + ... \rangle \\
= & \ a + \frac{F^{(4)}(0)}{4!} \langle u^4 \rangle + ... 
=  a + \frac{F^{(4)}(0)}{4!} \ \kappa^{l}_4 \ d^4 + O(d^6) \\ 
\approx & \ a + b \ d^4 \label{eq:fhat} 
\end{align}
for some constants, $a$ and $b$, showing that the optimization function has a fourth order dependency on $d$. For asymmetric distributions, we have
\begin{align}
\langle F(u) \rangle \approx & \ a + b \ d^3 \label{eq:fhatasym} 
\end{align}

Importantly, these equations enable us to formally reduce the gradient descent to one dimension. Let $\bw_j = e_j$ be the closest hidden feature to the initial weights, such that the overlap is given by the $j$-th component of the weights, $d = w_j$, and we assume $w_j > 0$ without loss of generality. We have that the gradient for the overlap has a supralinear dependency at small overlaps, with
\begin{equation}
\langle \hat{F}(d) \rangle \propto a + d^4 \implies  \frac{\partial \langle \hat{F}(d) \rangle}{\partial d} \propto d^3
\end{equation}
for symmetric distributions, and
\begin{equation}
\langle \hat{F}(d) \rangle \propto a + d^3 \implies  \frac{\partial \langle \hat{F}(d) \rangle}{\partial d} \propto d^2
\end{equation}
for asymmetric distributions.

\subsection{Learning time for stochastic gradients}

We calculate the optimal learning rate for small overlaps. After one update, with normalization, we have
\begin{align}
\bw_{t+1} = \frac{\bw_t + \eta \mathbf{g_t}}{||\bw_t + \eta \mathbf{g_t}||^2} 
\label{eq:sgdstep} 
\end{align}
Considering the component $d$ in the direction of the hidden feature $\bw_d$, we have
\begin{align}
d_{t+1} & \approx d_t + \eta \mu - \frac{1}{2} \eta^2 N \sigma^2 d_t \\
\implies  \Delta d & \approx \eta \mu - \frac{1}{2} \eta^2 N \sigma^2 d
\label{eq:sgdstepd} 
\end{align}
where we used that $||\bw_t||^2 = 1$, and $\mu = \langle \bw_d^T \mathbf{g_t} \rangle$,  $\sigma^2 = \langle (\bw_d^T \mathbf{g_t})^2 \rangle$. Optimizing the resulting overlap after a learning step, we have 
\begin{align}
\eta^* = argmax_\eta \langle \Delta d \rangle =  \frac{2 \mu}{N \sigma^2 d} 
\label{eq:etaopt} 
\end{align}
To estimate the learning time, we consider the mean dynamics on the component $d$, starting from $d_0$, for small overlaps $d \to 0$,
\begin{align}
T \approx \int_{d_0} \frac{1}{\Delta d} \partial d \ \propto \ N \int_{d_0} \frac{d}{\mu^2} \partial d
\label{eq:timedyn} 
\end{align}
where we have kept only variables that scale with $N$. We arrive at the scaling laws for the learning time, for the symmetric case, with K hidden features, 
\begin{align}
T & \propto N \int_{d_0} \frac{d}{d^6} \partial d \ \propto \ N [d^{-4}]_{d= \sqrt{log(K)/N}} \\
\implies T & \propto \frac{N^3}{log(K)^2}
\label{eq:timesym} 
\end{align}
and for the asymmetric case,
\begin{align}
T \propto \frac{N^2}{log(K)}
\label{eq:timeasym} 
\end{align}


\nolinenumbers



\printbibliography

@article{brito_nonlinear_2016,
	title = {Nonlinear {Hebbian} {Learning} as a {Unifying} {Principle} in {Receptive} {Field} {Formation}},
	volume = {12},
	issn = {1553-7358},
	url = {https://journals.plos.org/ploscompbiol/article?id=10.1371/journal.pcbi.1005070},
	doi = {10.1371/journal.pcbi.1005070},
	language = {en},
	number = {9},
	urldate = {2021-02-16},
	journal = {PLOS Computational Biology},
	author = {Brito, Carlos S. N. and Gerstner, Wulfram},
	month = sep,
	year = {2016},
	note = {Publisher: Public Library of Science},
	pages = {e1005070},
}

@inproceedings{le_building_2013,
	title = {Building high-level features using large scale unsupervised learning},
	url = {http://ieeexplore.ieee.org/xpls/abs_all.jsp?arnumber=6639343},
	urldate = {2015-12-21},
	booktitle = {Acoustics, {Speech} and {Signal} {Processing} ({ICASSP}), 2013 {IEEE} {International} {Conference} on},
	publisher = {IEEE},
	author = {Le, Quoc V.},
	year = {2013},
	note = {00673},
	pages = {8595--8598},
}

@inproceedings{krizhevsky_imagenet_2012,
	title = {Imagenet classification with deep convolutional neural networks},
	url = {http://papers.nips.cc/paper/4824-imagenet-classification-w},
	urldate = {2015-12-14},
	booktitle = {Advances in neural information processing systems},
	author = {Krizhevsky, Alex and Sutskever, Ilya and Hinton, Geoffrey E.},
	year = {2012},
	note = {03209},
	pages = {1097--1105},
}

@article{lecun_deep_2015,
	title = {Deep learning},
	volume = {521},
	copyright = {© 2015 Nature Publishing Group, a division of Macmillan Publishers Limited. All Rights Reserved.},
	issn = {0028-0836},
	url = {http://www.nature.com/nature/journal/v521/n7553/full/nature14539.html},
	doi = {10.1038/nature14539},
	language = {en},
	number = {7553},
	urldate = {2015-12-14},
	journal = {Nature},
	author = {LeCun, Yann and Bengio, Yoshua and Hinton, Geoffrey},
	month = may,
	year = {2015},
	note = {00047},
	pages = {436--444},
}

@article{berardi_critical_2000,
	title = {Critical periods during sensory development},
	volume = {10},
	url = {http://www.sciencedirect.com/science/article/pii/S0959438899000471},
	number = {1},
	urldate = {2015-12-14},
	journal = {Current opinion in neurobiology},
	author = {Berardi, Nicoletta and Pizzorusso, Tommaso and Maffei, Lamberto},
	year = {2000},
	note = {00388},
	pages = {138--145},
}

@article{rivera-alba_wiring_2011,
	title = {Wiring economy and volume exclusion determine neuronal placement in the {Drosophila} brain},
	volume = {21},
	url = {http://www.sciencedirect.com/science/article/pii/S0960982211011468},
	number = {23},
	urldate = {2015-12-14},
	journal = {Current Biology},
	author = {Rivera-Alba, Marta and Vitaladevuni, Shiv N. and Mishchenko, Yuriy and Lu, Zhiyuan and Takemura, Shin-ya and Scheffer, Lou and Meinertzhagen, Ian A. and Chklovskii, Dmitri B. and de Polavieja, Gonzalo G.},
	year = {2011},
	note = {00059},
	pages = {2000--2005},
}

@article{li_concise_2011,
	title = {Concise formulas for the area and volume of a hyperspherical cap},
	volume = {4},
	url = {http://docsdrive.com/pdfs/ansinet/ajms/0000/22275-22275.pdf},
	number = {1},
	urldate = {2015-12-14},
	journal = {Asian Journal of Mathematics and Statistics},
	author = {Li, Shengqiao},
	year = {2011},
	note = {00085},
	pages = {66--70},
}

@book{david_order_1970,
	title = {Order statistics},
	url = {http://onlinelibrary.wiley.com/doi/10.1002/0471667196.ess6023.pub2/pdf},
	urldate = {2015-12-14},
	publisher = {Wiley Online Library},
	author = {David, Herbert Aron and Nagaraja, Haikady Navada},
	year = {1970},
	note = {06572},
}

@article{choromanska_loss_2014,
	title = {The loss surface of multilayer networks},
	url = {http://arxiv.org/abs/1412.0233},
	urldate = {2015-12-13},
	journal = {arXiv preprint arXiv:1412.0233},
	author = {Choromanska, Anna and Henaff, Mikael and Mathieu, Michael and Arous, Gérard Ben and LeCun, Yann},
	year = {2014},
	note = {00022},
}

@article{mnih_playing_2013,
	title = {Playing atari with deep reinforcement learning},
	url = {http://arxiv.org/abs/1312.5602},
	urldate = {2015-12-13},
	journal = {arXiv preprint arXiv:1312.5602},
	author = {Mnih, Volodymyr and Kavukcuoglu, Koray and Silver, David and Graves, Alex and Antonoglou, Ioannis and Wierstra, Daan and Riedmiller, Martin},
	year = {2013},
	note = {00000},
}

@article{bengio_representation_2013,
	title = {Representation learning: {A} review and new perspectives},
	volume = {35},
	shorttitle = {Representation learning},
	url = {http://ieeexplore.ieee.org/xpls/abs_all.jsp?arnumber=6472238},
	number = {8},
	urldate = {2015-12-13},
	journal = {Pattern Analysis and Machine Intelligence, IEEE Transactions on},
	author = {Bengio, Yoshua and Courville, Aaron and Vincent, Pierre},
	year = {2013},
	note = {00000},
	pages = {1798--1828},
}

@article{cai_distributions_2013,
	title = {Distributions of {Angles} in {Random} {Packing} on {Spheres}},
	url = {http://arxiv.org/abs/1306.0256},
	urldate = {2015-06-22},
	journal = {arXiv:1306.0256 [math, stat]},
	author = {Cai, Tony and Fan, Jianqing and Jiang, Tiefeng},
	month = jun,
	year = {2013},
	note = {arXiv: 1306.0256},
}

@incollection{dauphin_identifying_2014,
	title = {Identifying and attacking the saddle point problem in high-dimensional non-convex optimization},
	url = {http://papers.nips.cc/paper/5486-identifying-and-attacking-the-saddle-point-problem-in-high-dimensional-non-convex-optimization.pdf},
	urldate = {2015-04-22},
	booktitle = {Advances in {Neural} {Information} {Processing} {Systems} 27},
	publisher = {Curran Associates, Inc.},
	author = {Dauphin, Yann N and Pascanu, Razvan and Gulcehre, Caglar and Cho, Kyunghyun and Ganguli, Surya and Bengio, Yoshua},
	editor = {Ghahramani, Z. and Welling, M. and Cortes, C. and Lawrence, N. D. and Weinberger, K. Q.},
	year = {2014},
	pages = {2933--2941},
}

@article{oja_learning_1991,
	title = {Learning in nonlinear constrained {Hebbian} networks.},
	journal = {Artificial Neural Networks},
	author = {Oja, E. and Ogawa, H. and Wangviwattana, J.},
	year = {1991},
	pages = {385--390},
}

@article{saxe_exact_2013,
	title = {Exact solutions to the nonlinear dynamics of learning in deep linear neural networks},
	url = {http://arxiv.org/abs/1312.6120},
	urldate = {2014-02-02},
	journal = {arXiv preprint arXiv:1312.6120},
	author = {Saxe, Andrew M. and McClelland, James L. and Ganguli, Surya},
	year = {2013},
}

@book{kandel_principles_2000,
	title = {Principles of neural science},
	publisher = {Appleton \& Lange},
	author = {Kandel, E. R. and Schwartz, J. H. and Jessell, T. M.},
	year = {2000},
}

@article{pfister_triplets_2006,
	title = {Triplets of {Spikes} in a {Model} of {Spike} {Timing}-{Dependent} {Plasticity}},
	volume = {26},
	number = {38},
	journal = {The Journal of Neuroscience},
	author = {Pfister, J. P and Gerstner, W.},
	year = {2006},
	pages = {9673--9682},
}

\end{document}